\documentclass[10pt,conference,letterpaper]{IEEEtran}

\usepackage{times}
\usepackage{subfigure}
\usepackage{graphicx}
\usepackage{color}
\usepackage{endnotes}
\usepackage{listings}
\usepackage{makecell}
\usepackage{multirow}
\usepackage[hyphens]{url}
\usepackage{cite}
\usepackage[colorlinks=true,linkcolor=black,citecolor=black,pdfborder={0 0 0}]{hyperref}
\usepackage{breakurl}
\usepackage{enumitem}
\usepackage{xcolor}
\usepackage{xspace}
\usepackage{microtype}
\usepackage[font=small,labelfont=bf]{caption}
\usepackage{comment}
\usepackage{textcomp}
\usepackage{algorithm}
\usepackage[noend]{algpseudocode}

\usepackage[english]{babel}

\usepackage{pifont}

\newcommand{\cmark}{\ding{51}}
\renewcommand{\paragraph}[1]{\smallskip\noindent {\bf #1}}

\newcommand\sysname{\textsc{Rodman}\xspace}

\begin{document}

\title{\huge Robust Data Preprocessing for Machine-Learning-Based Disk Failure
Prediction in Cloud Production Environments}

\author{Shujie Han$^\dagger$, Jun Wu$^\ddagger$, Erci Xu$^\ddagger$, Cheng
He$^\ddagger$, Patrick P. C. Lee$^\dagger$, \\
Yi Qiang$^\ddagger$, Qixing
Zheng$^\ddagger$, Tao Huang$^\ddagger$, Zixi Huang$^\ddagger$, Rui
Li$^\ddagger$\\
$^\dagger${\em The Chinese University of Hong Kong} \ \
$^\ddagger${\em Alibaba} }

\maketitle 

\begin{abstract}
To provide proactive fault tolerance for modern cloud data centers, extensive
studies have proposed machine learning (ML) approaches to predict imminent
disk failures for early remedy and evaluated their approaches directly on
public datasets (e.g., Backblaze SMART logs).  However, in real-world
production environments, the data quality is imperfect (e.g., inaccurate
labeling, missing data samples, and complex failure types), thereby degrading
the prediction accuracy.  We present \sysname, a robust data preprocessing
pipeline that refines data samples before feeding them into ML models.  We
start with a large-scale trace-driven study of over three million disks from
Alibaba Cloud's data centers, and motivate the practical challenges in ML-based
disk failure prediction.  We then design \sysname with three data preprocessing
techniques, namely failure-type filtering, spline-based data filling, and
automated pre-failure backtracking, that are applicable for general ML models.
Evaluation on both the Alibaba and Backblaze datasets shows that \sysname
improves the prediction accuracy compared to without data preprocessing under
various settings. 
\end{abstract}

\section{Introduction}
\label{sec:introduction}

Modern cloud data centers are equipped with millions of hard disk drives
spanning across the globe \cite{awsScale, xu2018}.  With such a
million-scale fleet of disks, disk failures become prevalent, complex, and
cascading.  For a data center equipped with high-end disks with an annual
failure rate of 1\% \cite{schroeder2007}, administrators need to handle
hundreds of disk failures/replacements on a daily basis.  While data centers
in practice are protected by RAID or erasure coding for fault tolerance,
frequent disk failures or concurrent disk failures in a single disk group
still lead to data unavailability \cite{ma2015}. 
To maintain data availability, it is crucial for administrators to be not
only reactive to disk failures that have occurred, but also {\em proactive} to
imminent disk failures, in large-scale disk deployment. 


Recent studies show that machine learning (ML) can accurately predict imminent
disk failures based on the already available SMART (Self-Monitoring, Analysis
and Reporting Technology) logs in modern disks
\cite{mahdisoltani2017,botezatu2016,li2014,li2016being,zhu2013,zhao2010,xu2018,hamerly2001,xiao2018}.
At a high level, existing ML-based disk failure prediction approaches first
build a classifier model using the features (e.g., SMART attributes) of a
training dataset that labels both positive and negative samples
(i.e., failed and healthy disks, respectively).  They then predict if a
disk will fail within days or weeks by feeding its features into the
classifier model.  Despite the extensive studies in the literature, applying
ML to disk failure prediction still faces three practical challenges in a
complex cloud production environment.
 



{\em (i) Inaccurate labeling.} Existing studies often label fail-stop events
as the only positive samples \cite{botezatu2016,xu2018,mahdisoltani2017}.  
However, field studies report that soon-to-fail disks experience latent
anomalies prior to actual failures, such as latent sector errors
\cite{bairavasundaram2007} and fail-slow symptoms \cite{gunawi2018}.  
Such pre-failure anomalies should also be labeled as positive samples, so
as to reflect more accurately the failure patterns.  To include
pre-failure anomalies, a na\"ive solution is that during the training phase,
we label the samples as positive if they appear within some pre-failure period
before the actual failures occur.  However, such a solution has two issues.
First, it is unclear how to configure the ``right'' length of the
pre-failure period without holistic measurements.  Second, the pre-failure
status of a failed disk is determined only after its failure actually occurs.
If the failure event occurs right after the training phase, the samples before
the failure event remain to be marked as healthy (negative), which still
inaccurately reflects the failure status of those samples. 

{\em (ii) Incomplete datasets.}  Disk failure prediction becomes inaccurate 
if it operates on incomplete SMART logs; for example, field studies report
that more than half of SMART failure signals are missing in failed disks
\cite{pinheiro2007}.  In production, disk monitoring systems may
stop recording failure signals due to network failures, software
maintenance/upgrades, system crashes, and human mistakes \cite{gunawi2016}.
In our production experience, some special commercial activities may require
the suspension of the disk monitoring systems for server offloading. 
One solution is to interpolate the missing failure signals, but improper
interpolation can severely compromise the prediction accuracy.  


{\em (iii) Diverse failure types.} Disk failures can manifest in different
forms, such as disk crashes, or fine-grained errors like latent sector
errors \cite{bairavasundaram2007,schroeder2010} and data corruptions
\cite{bairavasundaram2008}.  Some failures are due to accumulated
factors such as frequent occurrences of sector errors \cite{schroeder2010},
while others are due to transient factors such as power faults
\cite{zheng2013} or faulty interconnects \cite{jiang2008}.  SMART logs provide
insufficient details to address the diverse types of disk failures. 


We present \sysname, a \underline{ro}bust \underline{d}isk failure prediction
\underline{man}agement pipeline designed to address the above three
challenges.  Our goal is to emphasize the necessity of proper preprocessing on
the training datasets when deploying disk failure prediction in the field.
Note that some existing ML-based disk failure prediction approaches (e.g.,
\cite{botezatu2016,mahdisoltani2017,xiao2018}) are directly evaluated on the
public Backblaze SMART logs \cite{backblaze}. 
However, we find that the Backblaze dataset also shares the above three
challenges (i.e., only fail-stop events are labeled as positive samples, data
missing exists, and failure types are unavailable).  Such approaches, when
re-evaluated under different deployment settings, give inaccurate prediction
if the training datasets are not properly preprocessed. 


\sysname uses three data preprocessing techniques to refine data samples from
a large-scale dataset before feeding them into ML model training.  Its
techniques are applicable for general ML models.  \sysname is now deployed
in the production data centers at Alibaba Cloud.  We summarize our contributions
as follows.
\begin{itemize}[leftmargin=*]  
\item
We conduct a measurement study on a one-year span of disk logs from a fleet of
more than three million disks deployed at Alibaba Cloud, so as to motivate the
need of data preprocessing in practical disk failure prediction.  To our
knowledge, our measurement study is among the largest scale in the literature.
See \S\ref{sec:methodology} and \S\ref{sec:analysis}.
\item
We propose three data preprocessing techniques, namely: (i) {\em failure-type
filtering}, which selects only the positive samples of statistically
predictable failure types for training; (ii) {\em spline-based data filling},
which fills the values of missing samples via cubic spline interpolation
\cite{stoer2002} to account for any possible abrupt changes in such missing
samples; and (iii) {\em automated pre-failure backtracking}, which not only
automatically determines a window of pre-failure samples that are to be
labeled as positive for any soon-to-fail disk, but also avoids mis-labeling
any failed disk that appears right after the training phase.  See
\S\ref{sec:design}. 
\item
We evaluate the accuracy gain of each data preprocessing technique on both
Alibaba and Backblaze datasets.  \sysname achieves a disk failure prediction
rate of up to 92.8\% and 82.4\% under a false positive rate of 0.1\% and 4.0\%
on the Alibaba and Backblaze datasets, respectively, and improves the accuracy
by 6.2-23.7\% over state-of-the-art ML-based disk failure prediction approaches
\cite{botezatu2016,mahdisoltani2017}.  See \S\ref{sec:evaluation}. 
\end{itemize}

\section{Methodology}
\label{sec:methodology}



We collect data from a production cloud infrastructure at Alibaba Cloud that
covers multiple data centers across the globe.  Each data center is composed
of multiple nodes organized in racks, and each node is attached with multiple
disks.  

\begin{figure}[!t]
\centering
\includegraphics[width=3in]{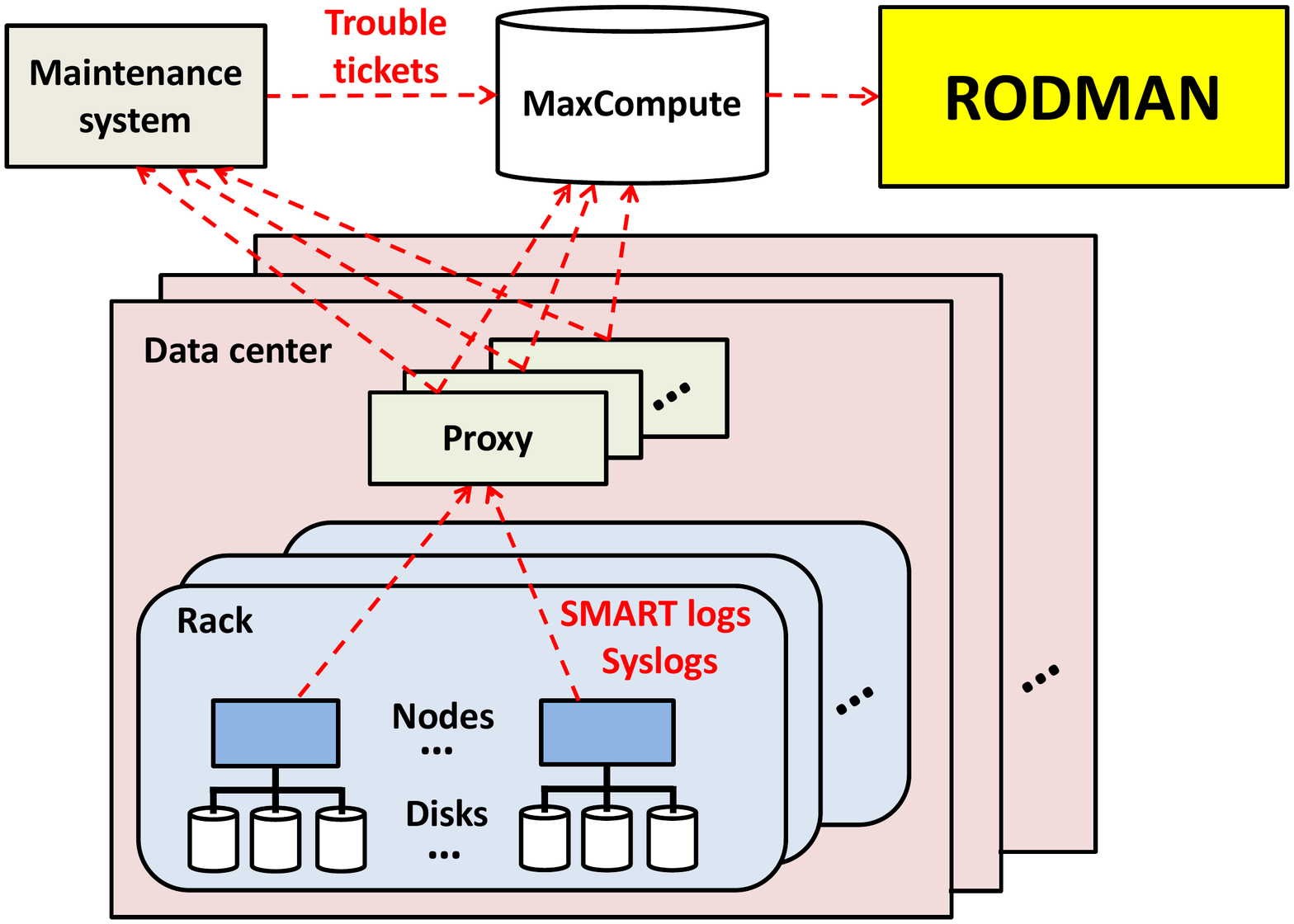}
\caption{Overview of data collection.}
\label{fig:collection}
\vspace{-6pt}
\end{figure}

Figure~\ref{fig:collection} depicts our data collection architecture.  
We collect three types of data: {\em SMART logs}, {\em system logs (syslogs)},
and {\em trouble tickets} (see details below).  Each node
collects SMART logs from its attached disks and syslogs from its operating
system, and periodically reports the collected data to one of the proxies of
our data collection architecture.  The proxies relay the collected data to a
data processing service, called MaxCompute \cite{maxcompute} that
supports an SQL-like interface for users to query
the stored data similar to Hive \cite{hive}.
Also, we deploy a maintenance system that identifies abnormal disk behaviors
and submits the corresponding trouble tickets to the data processing service.
\sysname takes the three types of data as input, performs data preprocessing,
and predicts disk failures. 


\begin{table*}[!t]
\centering
\begin{tabular}{|l|l|l|l|l|}
\hline
\multicolumn{1}{|c|}{\bf Source} & \multicolumn{1}{c|}{\bf Format} &
\multicolumn{1}{c|}{\bf Description} & \multicolumn{1}{c|}{\bf Granularity}  &
\multicolumn{1}{c|}{\bf Total size} \\ 
\hline\hline
SMART logs  & Key-value pairs &   Disk status reports  & Daily   & $>$ 1 billion log entries \\ 
\hline
Syslogs     & Plain texts     &   Kernel runtime events  & Hourly  & $>$ 4 trillion entries   \\ 
\hline
Trouble tickets &  Event entries &  Node ID, disk ID, timestamp,
and error messages & Per event & \texttildelow 31,000 events \\ 
\hline
\end{tabular}
\vspace{-3pt}
\caption{Dataset overview.}
\label{tab:overview}
\vspace{-6pt}
\end{table*}


We refer to our collected dataset as the {\em Alibaba dataset}. It was
collected over a one-year span from July 2017 to June 2018.  It covers a
population of over three million disks of tens of disk models from five disk
vendors, and the disks are deployed in dozens of data centers across the
globe.  Table~\ref{tab:overview} summarizes the dataset, which includes three
types of data. 

\paragraph{SMART logs.} SMART is a widely used disk status reporting tool that
collects performance and reliability statistics for different {\em attributes}
at disk firmware.  Our SMART logs cover the error-related or status-related
attributes for all disks at different times.  Each SMART log entry is
a key-value pair, in which the key is the disk ID (serial number),
and the value contains the disk model, disk vendor, timestamp, and a list of
SMART attributes.  Each attribute is a numerical value.  It is either
normalized (e.g., wearing percentage) or raw (e.g., power-on hours), 
and either instantaneous (e.g., temperature) or cumulative (e.g., number of
reallocated sectors).  

Our collected SMART attributes are vendor-specific and the number
of collected attributes varies across different disk vendors.  Nevertheless,
we find that most disk models report at least 38 attributes.  Also, all
attributes related to disk errors are universally reported at the daily
granularity.  In our one-year span, our SMART logs cover over one billion
entries. 


\paragraph{Syslogs.}  All nodes in this study run the major Linux
distributions.  The Linux kernel is installed with the syslog daemon
\cite{syslog} that is configured to collect system-level events at the hourly
granularity.  Each syslog event contains the process ID, timestamp, and
description of the event (e.g., the disk is not found). It also specifies the
logical partition (e.g., /dev/sda) that allows us to automatically correlate
each event with a disk.  Syslog events can be categorized into different
severity levels. We focus on five of them triggered by disk failures:
Emergency, Alert, Critical, Error, and Warning.  In our one-year span, our
syslogs cover over four trillion entries. 
	

\paragraph{Trouble tickets.}  Our maintenance system applies rule-based
checking procedures defined by administrators to the SMART logs and syslogs,
so as to check for any abnormal behaviors, including kernel-level
errors/alerts and SMART error-related attributes that are above some
pre-configured thresholds.  For any abnormal behavior, the maintenance system
generates a trouble ticket, which contains the node ID, disk ID, timestamp,
and a description about the error (e.g., a snippet of the syslog event
message).  In our one-year span, we have collected around 31,000 trouble
tickets. 


%
%

\section{Data Analysis}
\label{sec:analysis}



To motivate the design of \sysname, we first analyze the Alibaba dataset, with
emphasis on the disk failure patterns in production and the data missing issue
in the dataset (\S\ref{subsec:bigcloud}).  Due to privacy concerns, we cannot
publicize the Alibaba dataset at the time of the writing.  Thus, we also
analyze the public Backblaze dataset \cite{backblaze} for cross-validation 
(\S\ref{subsec:backblaze}). 

\subsection{Alibaba Dataset}
\label{subsec:bigcloud}


\paragraph{Annualized failure rates.}
We first estimate the annualized failure rates (AFRs) of several representative
disk models.  We define the AFR as the ratio of the number of failed disks
reported in our trouble tickets in the one-year span of our dataset to the
total number of disks.  

Recall that the Alibaba dataset covers tens of disk models from five disk
vendors (\S\ref{sec:methodology}).  We focus on the three largest models with
the most disks within each vendor (i.e., 15 in total).  We denote each disk
model by ``Vendor''``$k$'', where ``Vendor'' corresponds to a letter (`A',
`B', `C', `D', and `E') for each of the five vendors, and ``$k$'' (1 to 3)
corresponds to the $k$-th numerous model; for example, the disk model ``A2''
represents the second numerous model of vendor~A.   

Table~\ref{table:AFR} presents the results of the 15 selected disk models.
The AFR of each disk model is at most 3.8\%, and most disk models (10 out of
15) have an AFR of less than 1\%.  Such low AFRs are also found in previous
disk failure prediction studies \cite{botezatu2016,xu2018}, and hence we must
address the well-known {\em data imbalance} issue (i.e., the number of failed
disks is much less than that of healthy disks). 

\begin{table}[!t]
\centering
\begin{tabular}{|c|c|r@{.}l|r@{.}l|}
\hline
\textbf{Model} & \textbf{Vendor\%}  &
\multicolumn{2}{c|}{\textbf{Total\%}} & \multicolumn{2}{c|}{\textbf{AFR}}\\ \hline
\hline
A1  & 27.7\% & 12&5\% & 0&56\% \\
\hline
A2   & 21.1\% & 9&5\% & 0&67\% \\
\hline
A3  & 18.2\% & 8&2\% & 0&59\% \\
\hline
B1  & 29.2\% & 10&8\% & 0&68\% \\
\hline
B2  & 28.1\% & 10&3\% & 0&64\% \\
\hline
B3  & 20.1\% & 7&7\% & 0&38\% \\
\hline
C1  & 30.1\% & 2&5\% & 3&8\% \\
\hline
C2  & 18.6\% & 1&5\% & 2&8\% \\
\hline
C3  & 11.0\% & 0&88\% & 1&1\% \\
\hline
D1  & 41.6\% & 2&7\% & 0&30\% \\
\hline
D2  & 25.1\% & 1&6\% & 1&0\% \\
\hline
D3  & 18.9\% & 1&2\% & 2&0\% \\
\hline
E1  & 62.9\% & 1&7\% & 0&53\% \\
\hline
E2  & 14.6\% & 0&39\% & 0&78\% \\
\hline
E3  & 13.5\% & 0&35\% & 0&65\% \\
\hline
\end{tabular}
\vspace{-3pt}
\caption{Failure patterns for different disk models in the Alibaba dataset,
including the percentage of disks in the same vendor (``Vendor\%''), the
percentage of disks in the whole disk population (``Total\%''), and the
annualized failure rate (``AFR''). Due to privacy concerns, we cannot report
the number of disks for each disk model.}
\label{table:AFR}
\end{table}

\paragraph{Failure types.}
We now correlate both syslogs and trouble tickets to identify the manifested
failure types in production.  We classify the disk failures reported in
trouble tickets into six major types, namely data corruptions, I/O request
errors, unhandled errors, disk-not-found errors, unhealthy disks, and file
system corruptions.  We focus on the two largest disk models (in terms of the
number of disks in our dataset), A1 and B1, as the representatives.
Table~\ref{tab:failureCauses} shows each failure type and the proportions of
all failures in two disk models A1 and B1.  

Data corruptions and I/O request errors are the most dominant failure types
(44.1\% and 19.1\% for A1, and 39.2\% and 45.5\% for B1, respectively).  Data
corruptions refer to the unrecoverable data loss (e.g., integrity check
errors) that cannot be directly detected by disks \cite{bairavasundaram2008},
while I/O request errors refer to the failed I/O requests (e.g., due to bad
sectors or transient disconnection) that are detectable by disks.   

Unhandled errors (18.4\% and 9.8\% for A1 and B1, respectively) occur when
failed disks return unknown error codes after receiving I/O requests from the
kernel.  For example, the kernel does not recognize an error
code defined by the disk vendor, or failed disks encounter unexpected errors
(e.g., on-chip memory corruption) that give meaningless error codes. 

\begin{table}[!t]
\centering
\renewcommand*{\arraystretch}{1.1}
\begin{tabular}{|l|p{1.6in}|c|c|}
\hline
\multicolumn{1}{|c|}{\bf Type}   & \multicolumn{1}{c|}{\bf Description} & {\bf A1} & {\bf B1}\\ 
\hline\hline
\makecell[tl]{Data\\ corruptions} &  Unrecoverable data loss (e.g., integrity
check errors)
  & 44.1\% & 39.2\% \\ 
\hline 
\makecell[tl]{I/O request\\ errors} &  Failed I/O requests (e.g., bad sectors,
transient disconnection) & 19.1\% & 45.5\%\\ 
\hline
\makecell[tl]{Unhandled\\ errors} & Unknown disk error codes (e.g., 
unrecognized error codes, unexpected corruption to error codes) &
18.4\% & 9.8\% \\ 
\hline
\makecell[tl]{Disk-not-\\found errors} & Disk component failures (e.g.,
faulty interconnects) & 6.8\% & 1.5\% \\ 
\hline
\makecell[tl]{Unhealthy\\ disks} & Soon-to-fail disks reported by rule-based
checking & 2.1\% & Nil \\ 
\hline
\makecell[tl]{File system\\ corruptions} & Unmountable file system (e.g.,
metadata corruption) & Nil & 0.19\% \\ 
\hline
\makecell[tl]{Others} & No further information & 9.1\% & 3.9\% \\ 
\hline
\end{tabular}
\vspace{-3pt}
\caption{Failure types for disk models A1 and B1.}
\label{tab:failureCauses}
\end{table}

Disk-not-found errors (6.8\% and 1.5\% for A1 and B1, respectively) refer to
the permanent disk component failures (e.g., faulty interconnects) that make
the whole disks detached from the kernel.  They differ from I/O request errors,
in which the disks remain attached. 

Unhealthy disks (2.1\% for A1, but not found for B1) are reported by our
internal rule-based checking tool based on SMART logs.  The tool periodically
checks several SMART attribute values against pre-specified thresholds, and
reports the disks that likely fail soon (e.g., too many sector errors). 

File system corruptions (0.19\% for B1, but not found for A1) refer to the
disks with unmountable file systems.  They are a rare failure type, as modern
file systems provide protection in metadata (e.g., error-correction codes in
inode tables) \cite{alagappan2016}.

Also, we cannot determine the failure types of small fractions of failed disks
(9.1\% and 3.9\% for A1 and B1, respectively) due to the lack of recorded
information. 


\paragraph{Data missing.}
Ideally, the SMART data is collected for all disks continuously on a daily
basis, yet our dataset contains incomplete SMART data in both failed and
healthy disks (\S\ref{sec:introduction}). Here, a disk is said to
have data missing on a day if it misses {\em all} collected SMART attributes
instead of some particular SMART attributes. To show the severity of data
missing, we compute the {\em data missing ratio (DMR)} as the total number of
missing days over the expected total number of occurrence days if no data
missing occurs.  For failed disks, we count the expected occurrence days from
when the disks first appear in the dataset until they fail; for healthy disks,
we only focus on those that appear at the beginning of the dataset, and hence
the expected number of occurrence days is one year. 

To examine the data missing behavior before a failed disk is reported in a
trouble ticket, we measure the missing gap for any failed disk that has data
missing between the last reported date of the SMART logs and the reported date
of the trouble ticket.  Here, we focus on the percentages of data-missing
failed disks with such missing gaps of at least consecutive 10 and 25 days.  

\begin{table}[!t]
\centering
\begin{tabular}{|c|c|c|c|c|}
\hline
\multirow{2}{*}{\textbf{Model}} & \multicolumn{3}{c|}{\textbf{Failed}} &
\textbf{Healthy} \\
\cline{2-5}
& \textbf{DMR}  & \textbf{10 days\%} & \textbf{25 days\%} &
\textbf{DMR}\\
\hline
\hline
A1  & 21.1\% & 43.5\% & 31.7\% & 6.3\%\\
\hline
A2  & 21.3\% & 33.6\% & 14.4\% & 6.6\%\\
\hline
A3  & 22.4\% & 51.0\% & 36.7\% & 7.3\%\\
\hline
B1  & 28.1\% & 23.4\% & 15.1\% & 6.8\%\\
\hline
B2  & 29.5\% & 15.7\% & 13.4\% & 5.8\%\\
\hline
B3  & 20.8\% & 16.2\% & 12.5\% & 8.4\%\\
\hline
C1  & 15.4\% & 73.7\% & 36.8\% & 7.4\%\\
\hline
C2  & 13.2\% & 36.8\% & 21.1\% & 6.1\%\\
\hline
C3  & 18.5\% & 40.0\% & 40.0\% & 6.1\%\\
\hline
D1  & 19.5\% & 33.3\% & 16.7\% & 6.0\%\\
\hline
D2  & 13.2\% & 22.2\% & 33.3\% & 6.4\%\\
\hline
D3  & 12.9\% & 50.0\% & 0\% & 6.1\%\\
\hline
E1  & 16.7\% & 56.7\% & 56.7\% & 16.3\%\\
\hline
E2  & 14.2\% & 0\% & 0\% & 15.6\%\\
\hline
E3  & 14.6\% & 20.0\% & 20.0\% & 9.7\%\\
\hline
\end{tabular}
\vspace{-3pt}
\caption{Data missing analysis for different disk models in the Alibaba
dataset, including the data missing ratios (DMRs) for failed and healthy
disks, as well as the percentages of data-missing failed disks that miss at
least consecutive 10 and 25 days of samples before being reported in trouble
tickets (``10 days\%'' and ``25 days\%'', respectively).}
\label{tab:missing}
\vspace{-6pt}
\end{table}

Table~\ref{tab:missing} presents the results of our 15 selected disk models.
Data missing of failed disks is more severe than that of healthy
disks.  For failed disks, the DMRs among 15 disk models are 12.9-29.5\%,
while for healthy disks, the DMRs are 5.8-16.3\%.  Also, the data-missing
failed disks tend to miss data for a long consecutive period of days before
being reported in trouble tickets.  For example, 9 out of 15 disk models have
at least 30\% of data-missing failed disks with at least 10 days of data
missing, while 8 out of 15 disk models have at least 20\% of data-missing
failed disks with at least 25 days of data missing, before the reported dates
of trouble tickets. 

\begin{figure}[!t]
\centering
\begin{tabular}{@{\ }cc}
\includegraphics[width=1.6in]{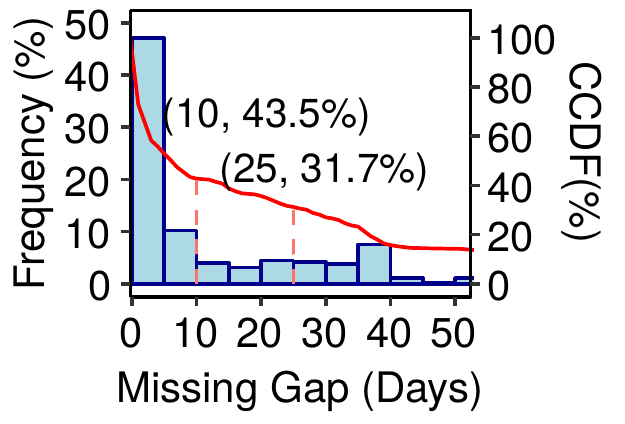} &
\includegraphics[width=1.6in]{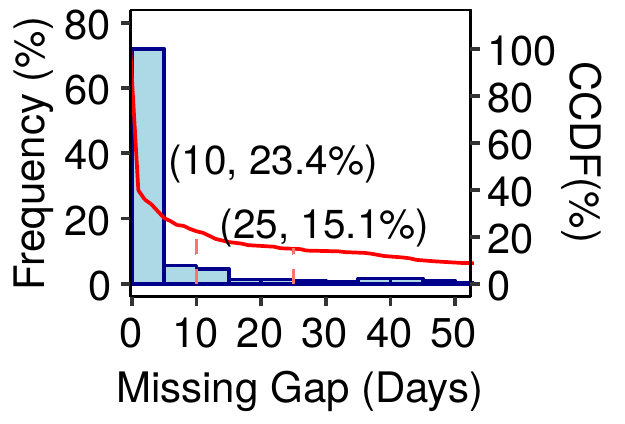} 
\vspace{-3pt}\\
\mbox{\small (a) A1} &
\mbox{\small (b) B1} 
\end{tabular}
\vspace{-3pt}
\caption{Missing gap distributions for data-missing failed disks in disk
models A1 and B1.}
\label{fig:missing}
\end{figure}

To understand the pre-failure data missing issue, we study the distribution of
the missing gap between the last recorded date of the SMART logs and the
recorded date of the trouble ticket for a data-missing failed disk in the two
largest disk models A1 and B1.  Figure~\ref{fig:missing} shows the histograms
and the complementary cumulative distribution functions (CCDFs) of the missing
gap distributions for both disk models.   Among
the data-missing failed disks, 43.5\% and 23.4\% of data-missing failed disks
in A1 and B1 have data missing for at least 10 consecutive days, respectively.
This suggests that failed disks may have already exhibited abnormal behaviors
before they are reported in trouble tickets, thereby leading to data missing. 

\subsection{Backblaze Dataset}
\label{subsec:backblaze}

We cross-validate our findings on the public Backblaze dataset
\cite{backblaze}. We focus on two disk models that are used in prior studies
\cite{botezatu2016,mahdisoltani2017,xiao2018} and have the largest population
and the most failures: (i) Hitachi HDS722020ALA330 (denoted by H1) over a
18-month span from April 2014 to September 2015, and (ii) Seagate ST4000DM000
(denoted by S1) over a 40-month span from September 2014 to December 2017.  

For H1, 133 out of 4,586 disks are failed, while for S1, 2,688 out of 32,320
disks are failed.  The corresponding AFRs for H1 and S1 are 1.9\% and 2.5\%,
respectively, similar to the AFRs in the Alibaba dataset.   We find that
data missing also exists in the Backblaze dataset (Table~\ref{tab:missing_bb}),
although the DMRs are much lower than those in the Alibaba dataset
(Table~\ref{tab:missing}).  Unfortunately, the Backblaze dataset does not
document the detailed failure symptoms or root causes, so we do not have
failure-type details.  

\begin{table}[t]
\centering
\begin{tabular}{|c|r@{.}l|c|c|c|}
\hline
\multirow{2}{*}{\textbf{Model}} & \multicolumn{4}{c|}{\textbf{Failed}} &
\textbf{Healthy} \\
\cline{2-6}
& \multicolumn{2}{c|}{\textbf{DMR}} & \textbf{10 days\%} & \textbf{25 days\%} &
\textbf{DMR}\\
\hline
\hline
H1 & 0&31\% & 0\% & 0\% & 0.20\% \\
\hline
S1 & 1&3\% & 0.08\% & 0.08\% & 0.36\% \\
\hline
\end{tabular}
\vspace{-3pt}
\caption{Data missing analysis for different disk models in the Backblaze
dataset.}
\label{tab:missing_bb}
\vspace{-6pt}
\end{table}

\section{\sysname Design}
\label{sec:design}

\begin{figure}[!t]
\centering
\includegraphics[width=3.2in]{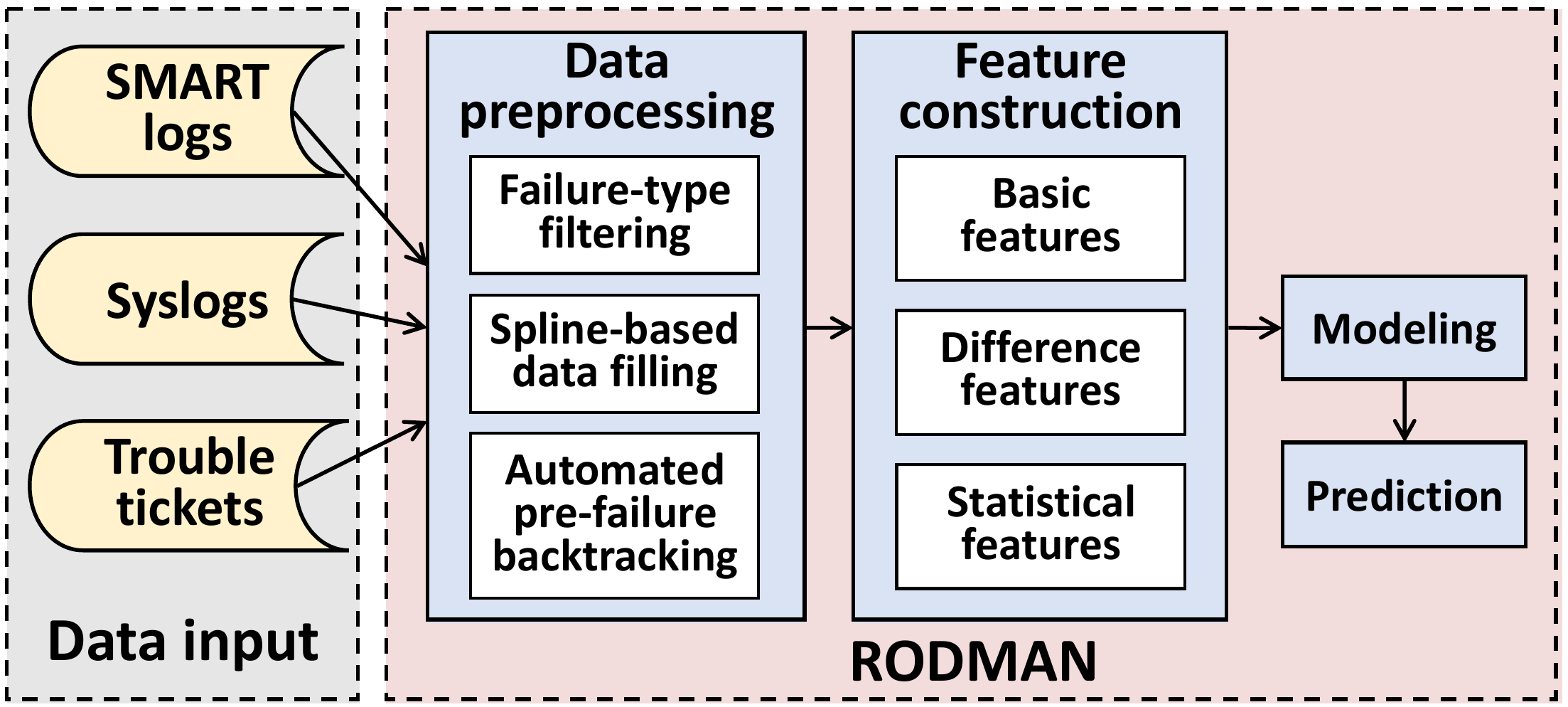}
\caption{Architecture of \sysname.}
\label{fig:rodman}
\end{figure}

We present \sysname, a robust disk failure prediction pipeline.
Figure~\ref{fig:rodman} shows \sysname's architecture.  \sysname takes the
SMART logs, syslogs, and trouble tickets as input. It converts SMART logs into
time-series samples, in daily granularity, for different disks. 
Each sample contains a set of SMART attributes at a time point. It uses
syslogs and trouble tickets for failure-type correlation and labeling,
respectively. It first refines the samples with three data preprocessing
techniques that are applied to training dataset before model training
(\S\ref{subsec:divFail}-\S\ref{subsec:pre-failure}).  It constructs features
for model training (\S\ref{subsec:construction}). It finally performs modeling
and outputs prediction results (\S\ref{subsec:modeling}).  We conclude this
section with the implementation details of \sysname (\S\ref{subsec:impl}). 

Our analysis focuses on two largest disk models of the Alibaba dataset, A1 and
B1, that have the most numbers of disks (and the most numbers of failed disks)
among all disk models we consider.  As they belong to different vendors, they
are expected to have varying failure characteristics
\cite{bairavasundaram2008,schroeder2010}.  We also validate our analysis for
other disk models from both Alibaba and Backblaze datasets
(\S\ref{sec:evaluation}).  \sysname focuses on offline learning and assumes
that the whole training dataset is available in advance, while we address
online learning on real-time data in future work.  



\subsection{Failure-Type Filtering}
\label{subsec:divFail}
  
Recall from Table~\ref{tab:failureCauses} that there are various types of
disk-related failures.  Our goal is to leverage the diverse failure types to
help our model training in \sysname. 


\paragraph{Analysis.}
Before examining the failure type information, we first analyze the
correlation between the SMART attributes and the disk failures reported
by our trouble tickets.  We compute Spearman's Rank Correlation Coefficient
(SRCC) \cite{spearman1987} between each of the SMART attributes and disk
failures.  SRCC ranges from -1 to +1 and shows the positive/negative
correlation between two variables; a larger absolute value (with the maximum
equal to one) implies that the two variables are more correlated, while 0
means no correlation. In our case, one variable refers to the raw value of a
SMART attribute, while another variable refers to the indicator variable
whether a disk is failed (1 means failed, or 0 otherwise). 

Table~\ref{tab:SMARTcorr} shows the four most failure-correlated SMART
attributes for disk models A1 and B1 with the largest absolute SRCC values.
Three of the SMART attributes appear in both A1 and B1, namely
SM5 (Reallocated Sectors Count), SM197 (Current Pending Sector Count),
and SM198 (Offline Uncorrectable Sector Count), and the remaining ones
are SM187 (Reported Uncorrectable Errors) for A1 and SM196 (Reallocated Event
Count) for B1.  We find that no SMART attribute in either disk model is
dominant, as the highest values are 0.401 and 0.352 for A1 and B1, respectively.
Furthermore, Table~\ref{tab:non-zero} shows the percentage of failed disks
with zero values for the four most failure-correlated SMART attributes for A1
and B1.  We find that 43.9\% and 78.2\% of failed disks have zero values in
all the four SMART attributes for A1 and B1, respectively, conforming to the
prior study that many failed disks have zero values in failure-correlated
SMART attributes \cite{pinheiro2007}.  Our results suggest that individual
SMART attributes are weak indicators of disk failures.  

\begin{table}[!t]
\centering
\begin{tabular}{|c|c|c|c|c|}
\hline
\textbf{Model} & \textbf{SM5}  & \textbf{SM187} & 
\textbf{SM197}  & \textbf{SM198} \\
\hline
A1 &  0.397  & 0.401 & 0.314 & 0.283\\
\hline  	
\hline
\textbf{Model} & \textbf{SM5}   & 
\textbf{SM196}  & \textbf{SM197}  & \textbf{SM198} \\
\hline  	
B1 & 0.352  & 0.325 &  0.336 & 0.330 \\
\hline  	
\end{tabular}
\vspace{-3pt}
\caption{Spearman's Rank Correlation Coefficients of the selected SMART
attributes and disk failures for A1 and B1.} 
\label{tab:SMARTcorr}
\end{table}
\begin{table}[!t]
\centering
\begin{tabular}{|c|c|c|c|c|c|}
\hline
\textbf{Model} & \textbf{SM5}  & \textbf{SM187} & 
\textbf{SM197}  & \textbf{SM198} & \textbf{All}\\
\hline
A1 &  49.6\%  & 81.9\% & 60.8\% & 60.9\% & 43.9\% \\
\hline  	
\hline
\textbf{Model} & \textbf{SM5}   & 
\textbf{SM196}  & \textbf{SM197}  & \textbf{SM198} & \textbf{All}\\
\hline  	
B1 & 92.1\%  & 92.1\% & 79.5\% & 94.0\% & 78.2\%\\
\hline  	
\end{tabular}
\vspace{-3pt}
\caption{Percentage of failed disks with zero values in the selected SMART
attributes for A1 and B1; ``All'' means that all four attributes have zero
values.} 
\label{tab:non-zero}
\end{table}
\begin{table}[!t]
\centering
\begin{tabular}{|c|c|c|c|c|c|c|c|c|}
\hline
\textbf{Model} & \textbf{SMART}& \textbf{1} & \textbf{2} & \textbf{3} & \textbf{4} &
\textbf{5} & \textbf{6} & \textbf{7} \\
\hline
\hline
\multirow{4}{*}{A1} & SM5 & \cmark & & \cmark & \cmark & \cmark &  &
\cmark\\
\cline{2-9}
& SM187 & \cmark & & & & \cmark & & \cmark \\
\cline{2-9}
& SM197 & \cmark & & & & \cmark & & \cmark\\
\cline{2-9}
& SM198 & \cmark &  & &\cmark &\cmark & & \cmark\\
\hline
\multirow{4}{*}{B1} & SM5 & \cmark & & & & & &\\
\cline{2-9}
& SM196 & & & & & & &\\
\cline{2-9}
& SM197 & & & & & & & \cmark \\
\cline{2-9}
& SM198 & \cmark & & & & & & \cmark\\
\hline
\end{tabular}
\vspace{-3pt}
\caption{KS test for evaluating the distribution differences of each SMART
attribute between failed and healthy disks for A1 and B1, grouped by failure
types (1: Data corruptions; 2: I/O request errors; 3: Unhandled errors; 4:
Disk-not-found errors; 5: Unhealthy disks; 6: File system corruptions; 7:
Others). A \cmark means that the distribution is statistically different.}
\label{tab:stat_compare}
\vspace{-6pt}
\end{table}

Nevertheless, by grouping the SMART attributes by failure types, we can
differentiate the distributions of SMART attribute values between the healthy
and failed disks.  Our insight is that only a subset of failures can be measured
by SMART logs.  If there is a larger discrepancy in the distributions, then
the trained classifier can recognize such failures more easily.  Specifically, 
we use the two-sample Kolmogorov-Smirnov (KS) test \cite{massey51} to
statistically evaluate the distribution discrepancies of each SMART attribute
between the healthy and failed disks.  Table~\ref{tab:stat_compare} shows the
KS test results, in which the statistically different distributions (with a
confidence level of 95\%) are marked with ticks (\cmark).  We call a failure
type {\em predictable} if there is more than one tick in the underlying
SMART attribute distributions.  Thus, the predictable failure types are data
corruptions, disk-not-found errors, and unhealthy disks for A1, and data
corruptions for B1.



We use the disk failures of predictable failure types to build our
classifier; on the other hand, we filter any disk failure of
unpredictable failure types.  We do not consider training a different model
for each failure type, as some failure types (e.g., unhealthy disks for A1 and
file system corruptions for B1) have very limited positive samples for
accurate training. 


\paragraph{Our training procedure.}
We summarize our training procedure based on failure-type filtering as
follows.  Before training a classifier for a particular disk model, we first
choose a configurable number (four in our case) of most failure-correlated
SMART attributes.  We then parse the syslogs to group the failure types and
compare the above SMART attribute distributions between the healthy and failed
disks for different failure types with the KS test.  Finally, we identify the
failure types as predictable if there is more than one failure-correlated
SMART attribute with statistically different distributions between the healthy
and failed disks.  When training the classifier, we only use the disk failures
of predictable failure types as positive samples and discard the
unpredictable ones.

\subsection{Spline-based Data Filling}

Before applying model training to a dataset, we must first address missing
data samples in every failed or healthy disk.  We discuss different data
filling approaches. 

\paragraph{Na\"ive data filling approaches.} One na\"ive approach is 
{\em forward filling}, which fills any missing sample with the value of any
last observed sample in the series (e.g., for the series ``1, 2, miss, 3, 4'',
we fill ``miss'' with ``2'').  Another na\"ive approach is {\em linear
interpolation}, which fits any missing sample on a linear line that connects
the endpoints of each missing gap (e.g., for the series ``1, 2, miss, 3, 4'',
we fill ``miss'' with ``2.5'').  Both approaches, albeit
implementation-friendly, unfortunately cannot address complicated changes in
SMART values. For example, the counts of error-related SMART attributes often
ramp up abruptly for a soon-to-fail disk. 

\begin{figure}[t]
\centering
\includegraphics[width=2.7in]{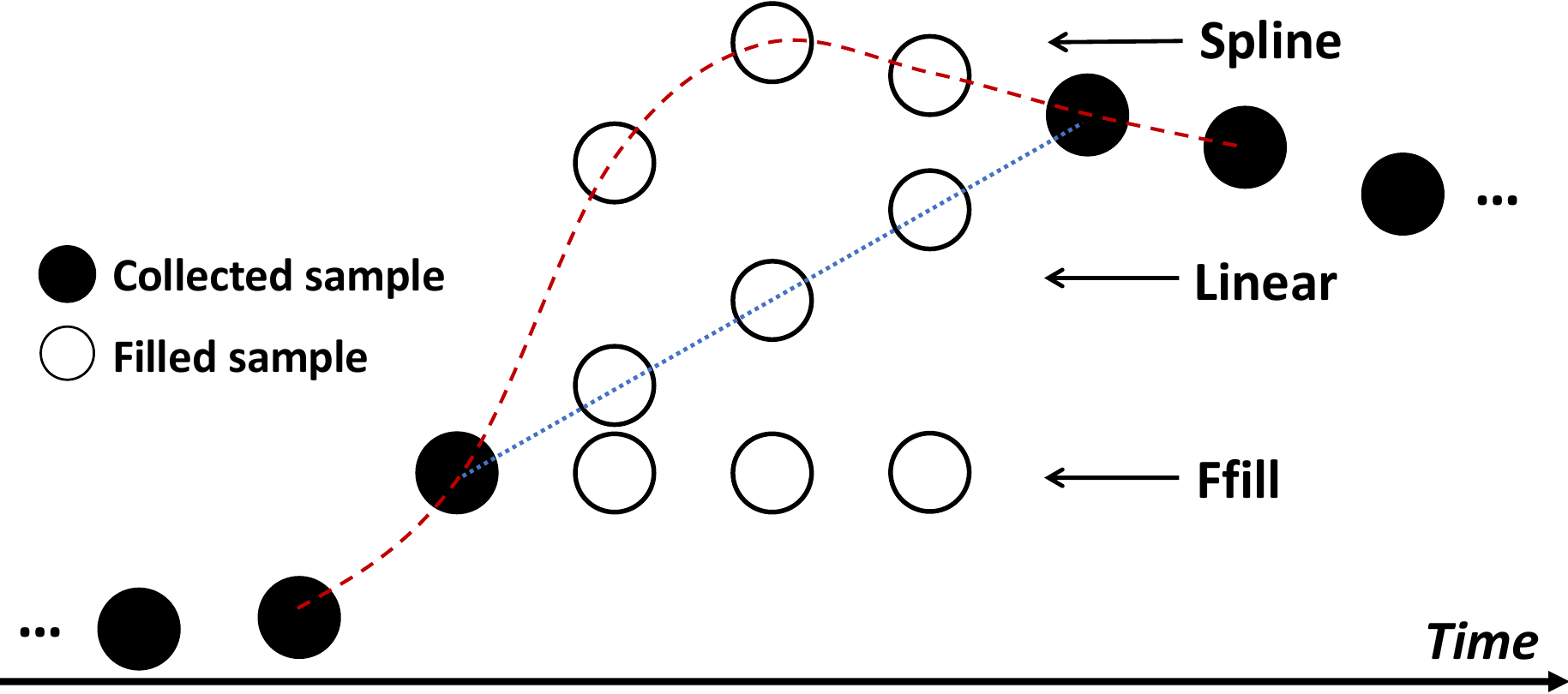}
\caption{Comparison of interpolation approaches: forward filling (Ffill),
linear interpolation (Linear), and cubic spline interpolation (Spline). Cubic
spline interpolation provides better fitting for abrupt changes than other two
approaches.}
\label{fig:interpolation}
\vspace{-6pt}
\end{figure}

\paragraph{Our data filling approach.}
\sysname uses {\em cubic spline interpolation} \cite{stoer2002} to
fill any missing samples.  At a high level, for each missing gap over a time
series, we select the two closest available data samples at both ends (i.e.,
four samples in total) to construct piecewise cubic polynomials that connect
all four samples while ensuring a ``smooth'' polynomial.  Cubic spline
interpolation provides better fitting for the abrupt changes than the na\"ive
approaches (see Figure~\ref{fig:interpolation} for comparisons), while
avoiding oscillated fitting that occurs in high-degree polynomial
interpolation (a.k.a. Runge's phenomenon).  It reduces to a linear function
for the time series with stable monotonic trends (e.g., power-on-hours).

We address two corner cases in data filling.  First,
the missing gap may be open-ended, such that the missing samples may be
observed at the very beginning (or at the very end) of the dataset. In this
case, we re-use the first (or the last) piecewise polynomial in the time
series from our interpolation results to extrapolate those missing samples.
Second, if the missing gap lasts longer than some pre-specified threshold
(currently set to 30 days), we drop all the samples for the time series. 

We do not claim that spline-based data filling can address the missing
samples for all data types, yet it improves the prediction accuracy over
na\"ive approaches (\S\ref{subsec:results1}) and suffices for our practical
deployment.  Optimizing the data filling strategy is our future work.

\subsection{Automated Pre-Failure Backtracking}
\label{subsec:pre-failure}

To accurately reflect the disk failure characteristics, we propose to mark the
pre-failure samples of failed disks as positive.  However, the length of the
pre-failure period is typically unknown.  Thus, we need to address two
questions: (i) How do we automatically determine the length of the pre-failure
period? (ii) Given the pre-failure period, how do we properly backtrack and
mark disk samples? 

\paragraph{Automated determination of the pre-failure period.} We identify the
pre-failure period using {\em Bayesian change point detection}
\cite{fearnhead2006} for failed disks in SMART logs, so as to pinpoint the
time when the SMART values of failed disks significantly vary (i.e., the
failure behaviors start to appear).  Specifically, during our training
procedure for each disk model, we first identify the four most
failure-correlated SMART attributes based on SRCC and select the failed disks
of predictable failure types (\S\ref{subsec:divFail}).  We pick a
sufficiently long detection window (e.g., 60 days in our implementation)
before the failure happens for each failed disk that has already been labeled
from trouble tickets.  We compute the change probability (i.e., the posterior
distribution of a time-series up to a sample given the time-series before the
sample)
of each of the four selected SMART attributes on each
day for each failed disk over the detection window.  Given a sequence of
change probabilities over the detection window of each failed disk, we use the
z-score (i.e., the number of standard deviations from the mean of
change probabilities) to measure if the change is significant.  We now choose
the critical z-score values as $\pm$2.5 of the standard deviation (i.e., a
confidence level of 98.76\%), such that if the z-score of the change
probability falls outside the range [-2.5, 2.5], we regard the change as
significant.  We obtain the numbers of days between the significant change and
the disk failure for all failed disks. Finally, we find the 75th-percentile of
the numbers of days obtained for each of our selected failure-correlated SMART
attributes, and choose the maximum 75-th percentile among the SMART attributes
as our pre-failure period. 

Based on our calculation, the pre-failure periods of A1 and B1 are 29 and 27
(days), respectively.  Note that finding the pre-failure period can be
automated, and the length of the pre-failure period varies for different disk
models.

\paragraph{Pre-failure backtracking.} Given the pre-failure period, we
backtrack the pre-failure samples of a failed disk and mark them as positive
samples.  We define a {\em backtracking window} as the number of backtracked
days in which we mark the samples of a failed disk as positive.  Currently, we
set the backtracking window length (denoted by $n$) as the pre-failure period
that is automatically selected for each disk model.

\begin{figure}[!t]
\centering
\begin{tabular}{@{\ }c}
\includegraphics[width=2.5in]{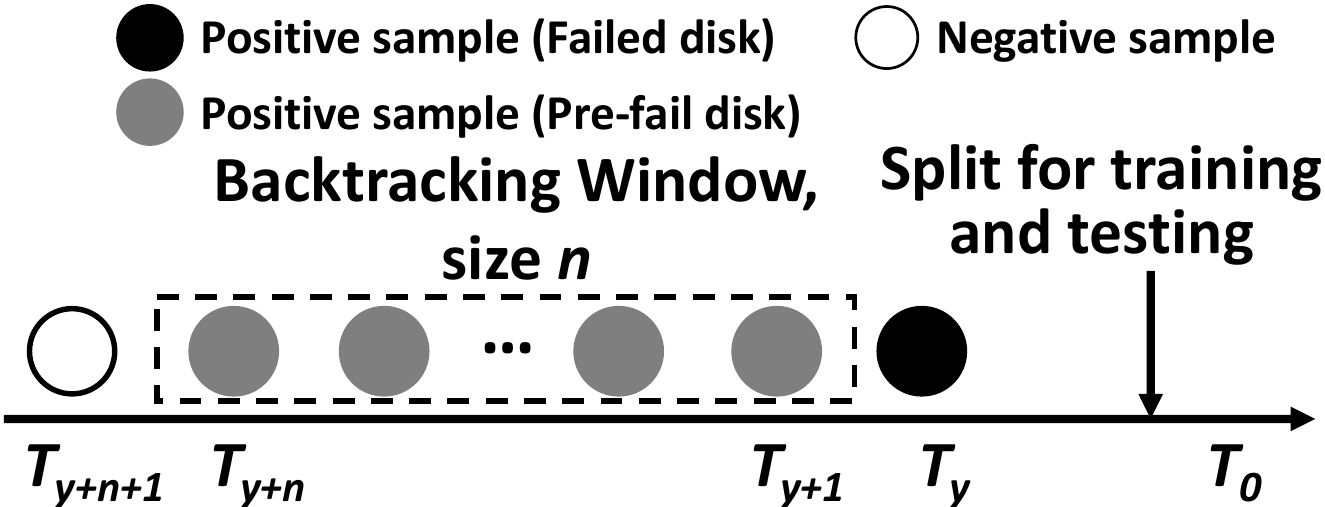}\\
\mbox{\small (a) Baseline pre-failure backtracking without observation window}
\vspace{6pt}\\
\includegraphics[width=3.2in]{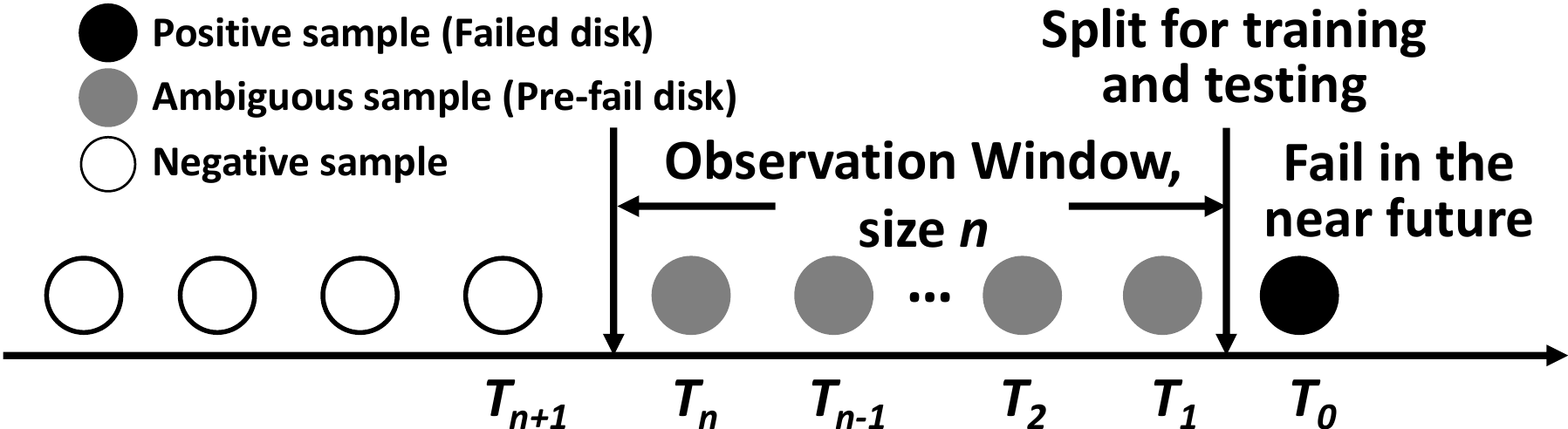}\\
\mbox{\small (b) Pre-failure backtracking with observation window}
\end{tabular}
\caption{Pre-failure backtracking.}
\label{fig:backtracking}
\vspace{-6pt}
\end{figure}

Figure~\ref{fig:backtracking}(a) shows the {\em baseline} design of our
pre-failure backtracking method.  Suppose that the training phase is set right
before day~$T_0$ and the testing phase starts on day~$T_0$.  Let $T_i$ be the
$i$-th backtracked day before $T_0$ at which a sample $s_i$ is observed, where
$i > 0$ (note that $T_i$ is earlier than $T_0$).  Let $T_y$ (for some $y > 0$)
be the last day on which the positive sample is observed (i.e., the disk
failure occurs).  Each dot in the figure represents a sample of SMART logs in
daily granularity.  We mark a failed disk at $T_y$ as a positive sample
(denoted by a black dot).  We also mark $n$ samples (denoted by gray dots)
prior to the disk failure as positive samples.  The earlier samples
before the backtracking window are marked as negative samples (i.e., healthy)
(denoted by white dots). 

However, there exists a potential issue with the baseline pre-failure
backtracking design.  Recall that the maintenance system generates the trouble
tickets of failed disks on the day of disk failure occurrence
(\S\ref{sec:methodology}).  Thus, we regard the disks that are not reported in
the trouble tickets as healthy in the training dataset, but the disks may fail
right after the training phase and their samples now become ``mis-labeled''
as negative.  Figure~\ref{fig:backtracking}(b) illustrates the issue.  If a disk
fails right after $T_0$ (i.e., the black dot), then we are supposed to mark
all samples of its backtracking window as positive samples (i.e., the gray
points). However, we do not have this information during model training as the
failure happens after the training phase.  From this example, we see that for
healthy disks, their samples in the backtracking window are actually ambiguous
samples, as they may fail right after the training phase. 

To avoid mislabeling negative samples, we propose an {\em observation window} 
to {\em drop} the ambiguous samples of pre-failure disks that are potentially
mislabelled as negative.  We now set the observation window as the
backtracking window $n$ and drop the (ambiguous) samples from $T_n$ to $T_1$.

\paragraph{Summary.}  Our complete pre-failure backtracking scheme with the
observation window works as follows.  If a disk is healthy, we drop all its
samples within the observation window (from $T_n$ to $T_1$) and mark any
sample outside the observation window (from $T_{n+1}$ backwards) as negative.
Otherwise, if a disk is failed (say in day $T_y$), we backtrack $n$ samples
within the backtracking window (i.e., from $T_{y+n}$ to $T_{y+1}$) to mark
them as positive, as well as the remaining samples outside the backtracking
window as negative.

\subsection{Feature Construction} 
\label{subsec:construction}

Our feature construction uses three categories of features.
\begin{itemize}[leftmargin=*] 
\item 
{\bf Basic features}: We select the error-related and status-related SMART
attributes (\S\ref{sec:methodology}) as the basic features. 
\item 
{\bf Difference features}: For each SMART attribute selected as a basic
feature, we calculate the difference values over two consecutive samples in
the time-series. 
\item 
{\bf Statistical features}: We compute six statistical characteristics for
each basic/difference feature over a window of samples: arithmetic mean,
standard deviation, median, exponential weighted moving average, sum, and
difference between the start and end points of the window.  We use two
window sizes, 7 and 14 days, for computing the above statistical
characteristics for each basic/difference feature. 
\end{itemize}

For example, for A1 (B1), we obtain 1,248 (988) features, including 48 (38)
basic features, 48 (38) difference features, and 1,152 (912) statistical
features.  Also, for H1 (S1) in the Backblaze dataset
(\S\ref{subsec:backblaze}), we obtain 884 (1,248) features, including 34 (48)
basic features, 34 (48) difference features, and 816 (1,152) statistical
features. 


\subsection{Modeling} 
\label{subsec:modeling}

We formulate our disk failure prediction problem as a time-series binary
classification problem. Among many classification algorithms, researchers
\cite{caruana2006,caruana2008} show that tree-based ensemble learning methods
(e.g., random forests and boosted trees) have the best overall prediction
accuracy.  As \sysname generates fewer than 4,000 features, it is shown that
boosted trees are preferred \cite{caruana2006,caruana2008}.
By default, \sysname uses LightGBM \cite{ke2017}, a boosted tree model with
high training performance and memory efficiency, yet \sysname is also
applicable for other ML models (\S\ref{subsec:results2}).  Note that we do not
claim the novelty of our modeling approach.

\subsection{Implementation}
\label{subsec:impl}

We implement the whole \sysname pipeline in Python with around 2,000 LoC.
We use the authors' Python implementation of LightGBM \cite{lightgbm}, and
realize other ML algorithms for comparisons using various Python tools (e.g.,
NumPy \cite{numpy}, SciPy \cite{scipy}, and Scikit-learn \cite{sklearn}).
\sysname is currently deployed at Alibaba Cloud. 

We also realize \sysname for the Backblaze dataset.  Since the Backblaze
dataset does not include failure types, we only realize automated pre-failure
backtracking and spline-based data filling. Our \sysname prototype is now
open-sourced (\S\ref{sec:introduction}). 





\section{Experiments}
\label{sec:evaluation}

We evaluate \sysname on both Alibaba and Backblaze datasets and show that
its data preprocessing improves the prediction accuracy.  We address
the following questions:
\begin{itemize}[leftmargin=*]  
\item 
Can automated pre-failure backtracking identify the proper backtracking window
length and improve the prediction accuracy?  (Exp\#1)
\item
Can failure-type filtering (i.e., using only predictable failure types for
training) accurately predict disk failures of both predictable and
unpredictable failure types? (Exp\#2)
\item 
How effective is spline-based data filling?  (Exp\#3)
\item 
Can \sysname improve the prediction accuracy over state-of-the-art disk failure
prediction approaches? (Exp\#4)
\item 
How does the prediction accuracy of \sysname vary for different configuration
parameters (e.g., false positive rates and training periods)?  (Exp\#5 and
Exp\#6)
\end{itemize}


\subsection{Evaluation Methodology}
\label{subsec:evalmeth}

\noindent
{\bf Configuration.}  For the Alibaba dataset, recall that it spans from July
2017 to June 2018 (\S\ref{subsec:bigcloud}).  By default, we set the training
phase from July 2017 to April 2018 (10 months) and the testing phase for May
2018 (one month).  \sysname predicts disk failures in the testing phase.  We
set a high training-to-testing ratio to 10:1 by default, so as to have
sufficient positive samples for training (as a reference, previous work sets
the training-to-testing ratio to 4:1 \cite{botezatu2016} or 3:1
\cite{mahdisoltani2017} for the Backblaze dataset evaluation); nevertheless,
we also consider smaller training-to-testing ratios starting from 2:1 (Exp\#6)
and in the Backblaze dataset evaluation (see below).  To mitigate
data imbalance and the bias toward the substantial volume of negative samples,
our training chooses the positive samples over the entire training phase,
while choosing the negative samples only on the last day observed (i.e.,
the end of April 2018 without the observation window, or the last day before
the observation window if enabled). 
		 
For the Backblaze dataset (\S\ref{subsec:backblaze}), we change the
training-to-testing ratio to 3:1 as in \cite{mahdisoltani2017}.  As in the
Alibaba dataset evaluation, our training chooses the positive samples over
the entire training phase and the negative samples only on the last day
observed to mitigate data imbalance. 
Given the long dataset durations (18 months for H1 and 40 months for S1), 
we apply model training in a sliding fashion to allow multiple runs.
Specifically, we use the first three months (e.g., January to March) as the
training phase and the fourth month (i.e., April) as the testing phase.  We
slide one month (i.e., February to April for training and May for testing) for
the next run, and so on.  We have 15 and 37 runs in total for H1 and S1,
respectively.  We present the average results over all runs.  


\paragraph{Metrics.}
Our disk failure prediction is a binary classification problem (i.e., failed
or healthy).  We focus on two metrics:
\begin{itemize}[leftmargin=*] \parskip=0pt \itemsep=0pt \topsep=0pt
\item 
\textbf{True positive rate (TPR)}: The ratio of the number of predicted
failed disks to the total number of actual disk failures in one-month testing. 
\item 
\textbf{False positive rate (FPR)}: The ratio of the number of falsely
predicted failed disks (which are indeed healthy) to the total number of
healthy disks in one-month testing.
\end{itemize} 

For the Alibaba dataset, we set the default FPR threshold to 0.1\% as in
Microsoft's production \cite{xu2018}.  We mainly present the results for disk
models A1 and B1, while we address other disk models in the Alibaba dataset in
\S\ref{subsec:results2}.  For the Backblaze dataset, we set a higher
default FPR threshold to 4\% to make the TPR comparable to that for the
Alibaba dataset.  Note that we still observe the accuracy gain of \sysname for
different FPR thresholds (Exp\#5).

\subsection{Analysis of Data Preprocessing Techniques}
\label{subsec:results1}


\begin{figure}[t]
\centering
\begin{tabular}{c}
\includegraphics[width=3in]{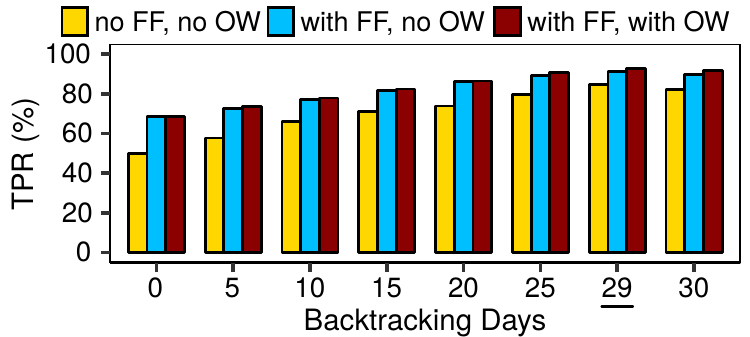}
\vspace{-3pt} \\
\mbox{\small (a) A1} 
\vspace{3pt} \\
\includegraphics[width=3in]{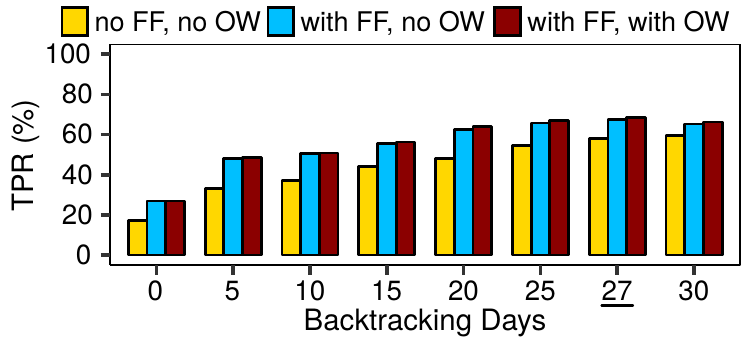} 
\vspace{-3pt} \\
\mbox{\small (b) B1}
\end{tabular}
\vspace{3pt} \\
\begin{tabular}{c@{\ }c}
\includegraphics[width=1.6in]{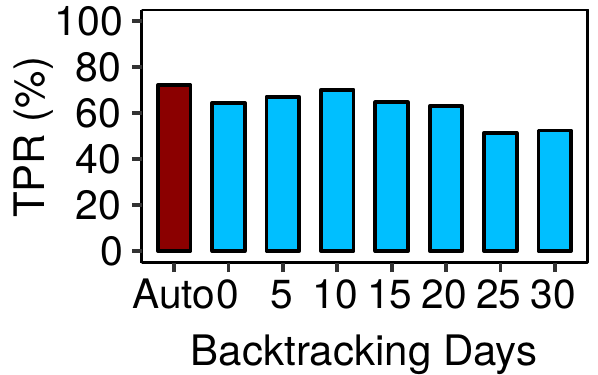} &
\includegraphics[width=1.6in]{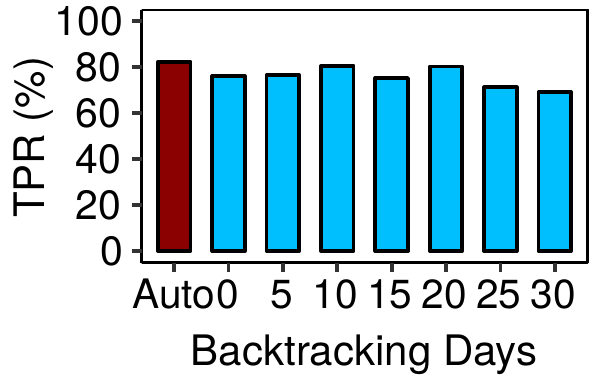}
\vspace{-3pt} \\
\mbox{\small (c) H1} &
\mbox{\small (d) S1}
\end{tabular}
\caption{Exp\#1 (Effectiveness of automated pre-failure backtracking).  We
show the TPRs of every five days (from 0 to 30 backtracking days) and the TPR
of our automatically determined backtracking days (29 days for A1 and 27 days
for B1 as underlined in the figures; ``Auto'' for H1 and S1).  For A1 and B1, 
we compare two components: failure-type filtering (``FF'') and observation
window (``OW'').} 
\label{fig:exp1}
\vspace{-6pt}
\end{figure}

\noindent
{\bf Exp\#1 (Effectiveness of automated pre-failure backtracking).}
We first consider the Alibaba dataset.  Recall that our automated pre-failure
backtracking selects the backtracking window as 29 and 27 days for A1 and B1,
respectively (\S\ref{subsec:pre-failure}).  We evaluate \sysname for
various lengths of backtracking windows and show the effectiveness of
automated pre-failure backtracking.  Before running the experiment, we have
enabled spline-based data filling.  We also compare two components of
\sysname: (i) failure-type filtering (i.e., using only the disk failures of
predictable failure types for training instead of using all disk failures),
and (ii) observation window (i.e., discarding a window of samples before the
end of training for healthy disks). 
 
Figures~\ref{fig:exp1}(a) and \ref{fig:exp1}(b) show the TPR results versus
the number of backtracking days for A1 and B1, respectively.  First, the TPR
increases as we increase the number of backtracking days from zero, as we now
introduce more positive samples to alleviate data imbalance.  Without
pre-failure backtracking (the zero-day case), \sysname can only correctly
predict 68.7\% of disk failures of A1 and 26.9\% disk failures of B1 when
both the failure-type filtering and the observation window are enabled.  As we
increase the number of backtracking days, the TPR for both disk models
increases and reaches the highest between the 25 and 30 backtracking days
(note that the TPR differences in this range are within 2\% for both disk
models).   In particular, the TPR reaches 92.8\% for A1 and 68.4\% for B1 in
our automatically determined backtracking days (29 and 27 days, respectively).
After 30 backtracking days, the TPR decreases, as we may now falsely inject
more positive samples.  This shows that our automated pre-failure backtracking
can reach the highest TPR range.

We next compare the TPR with and without failure-type filtering for A1 and B1.
With failure-type filtering, the TPR increases for both disk models.  For
example, without pre-failure backtracking (the zero-day case), failure-type
filtering itself increases the TPR from 49.8\% to 68.7\% for A1 and from
17.1\% to 26.9\% for B1.  We provide more detailed evaluation of failure-type
filtering in Exp\#2. 

We further study the TPR gains with the observation window (which equals
the number of backtracking days (\S\ref{subsec:pre-failure})).  Enabling the
observation window for non-zero backtracking days on top of failure-type
filtering further increases the TPR for both disk models in {\em all} cases, by
0.20-2.09\% and 0.48-1.77\% for A1 and B1, respectively.  Although the accuracy
improvement is small compared to the gains of failure-type filtering, we
regard the observation window as a complementary approach for pre-failure
backtracking: while pre-failure backtracking focuses on adding positive
samples, the observation window keeps only the obvious negative samples and
drops the ambiguous ones.  

Finally, we validate our analysis with the Backblaze dataset.  We study the
accuracy gain when automated pre-failure backtracking with the observation
window is enabled (recall that the Backblaze dataset has no failure types).  
Figures~\ref{fig:exp1}(c) and \ref{fig:exp1}(d) show the TPR results versus
the number of backtracking days for H1 and S1, respectively; we omit
the plots for the backtracking days from 35 to 60 days as the TPR keeps
decreasing.  The automated backtracking days vary from 2 to 58 days for H1 and
from 3 to 40 days for S1 across different runs.  On average, automated
pre-failure backtracking achieves the highest TPR, which is 72.3\% for H1 and
82.4\% for S1.

\paragraph{Exp\#2 (Effectiveness of failure-type filtering).} Recall that with
failure-type filtering, we use the failures of predictable failure types
(i.e., data corruptions, disk-not-found errors, and unhealthy disks for A1,
and data corruptions for B1) for training.  We now give a breakdown of the
prediction accuracy of \sysname for each failure type, with and without
failure-type filtering.  Here, we have enabled automated pre-failure
backtracking with the observation window and spline-based data filling. We
focus on A1 and B1 in the Alibaba dataset. 

Table~\ref{tab:exp2} shows the TPR results for different failure types with and
without failure-type filtering.  We make two observations. First, \sysname
successfully identifies the failures of unpredictable failure types (e.g.,
unhandled errors), even being trained with only predictable failure types.
Second, for most failure types, failure-type filtering improves the prediction
accuracy.  An exception is disk-not-found errors, which show a lower TPR for
both disk models with failure-type filtering (another exception is
``Others''), even though disk-not-found errors are a predictable failure type
in A1. 

\begin{table}[!t]
\centering
\begin{tabular}{|l|c|c|c|c|}
\hline
\multirow{2}{*}{\textbf{Type}} & \multicolumn{2}{c|}{\textbf{A1}} &
\multicolumn{2}{c|}{\textbf{B1}} \\
\cline{2-5}
& \textbf{w/o FF} & \textbf{w/ FF} & \textbf{w/o FF} & \textbf{w/ FF} \\
\hline
\hline
\makecell[l]{Data corruptions} & 81.4\% & 95.1\% & 51.0\% & 70.2\% \\
\hline
\makecell[l]{I/O request errors} & 88.5\% & 94.2\% & 56.8\% & 61.1\% \\
\hline
\makecell[l]{Unhandled errors} & 92.0\% & 96.0\% & 60.6\% & 69.7\% \\
\hline
\makecell[l]{Disk-not-found errors} & 92.3\% & 76.9\% & 50.0\% & 0\% \\
\hline
\makecell[l]{Unhealthy disks} & 42.9\% & 71.4\% & Nil & Nil \\
\hline
\makecell[l]{File system corruptions} & Nil & Nil & Nil & Nil \\
\hline
\makecell[l]{Others} & 95.5\% & 90.9\% & 100.0\% & 100.0\% \\
\hline
\end{tabular}
\vspace{-3pt}
\caption{Exp\#2 (Effectiveness of failure-type filtering). We report the
TPR for each failure type with or without failure-type filtering (``FF'');
``Nil'' means no failures belong to the failure type in testing.}
\label{tab:exp2}
\vspace{-6pt}
\end{table}

One explanation to the above observations is that various disk
failure symptoms can share the same root causes (e.g., both data corruptions
and I/O request errors can be caused by a failing disk sector)
\cite{gunawi2018}.  Thus, training a classifier with only representative
failure types can correctly identify different types of imminent disk
failures, including those of unpredictable failure types.  The
trade-off is that by filtering the unpredictable failure types, we may
unexpectedly drop disk failures whose root causes manifest as both predictable
and unpredictable failure types.  Nevertheless, the overall prediction
accuracy improves, as shown in Exp\#1.  



\paragraph{Exp\#3 (Effectiveness of spline-based data filling).}  We
evaluate different data filling approaches, including forward filling, linear
interpolation, and cubic spline interpolation.  Note that the data filling
approaches fill in different values, and hence return different
backtracking windows from our automated pre-failure backtracking.  For
example, for the Alibaba dataset, the backtracking windows for forward
filling, linear interpolation, and cubic spline interpolation are chosen as
26, 28, and 29 days, respectively for A1, as well as 24, 27, and 27 days,
respectively, for B1.  We also evaluate the no-filling case by training the
raw dataset directly, in which we set the backtracking windows as 29 and 27
days for A1 and B1, respectively, as in cubic spline interpolation.  Here, we
have enabled failure-type filtering for A1 and B1. 

\begin{figure}[t]
\centering
\includegraphics[width=2.8in]{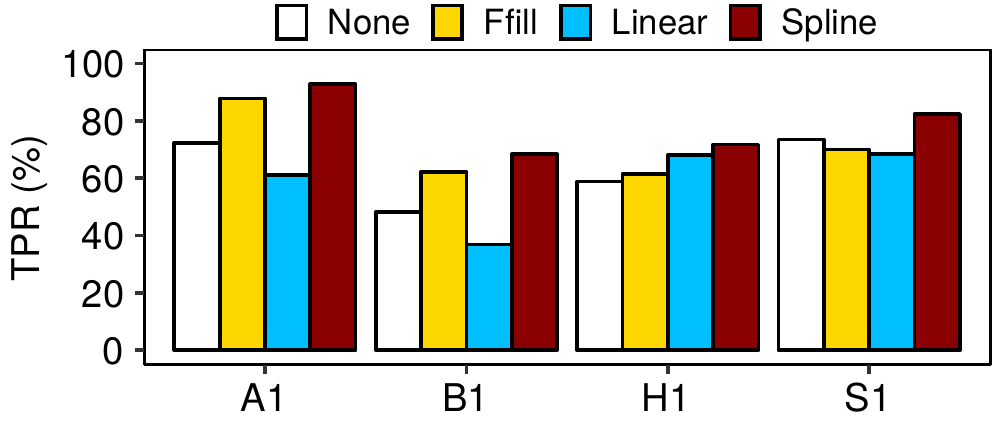}
\caption{Exp\#3 (Effectiveness of spline-based data filling). We compare
no-filling (``None''), forward filling (``Ffill''), linear interpolation
(``Linear''), and cubic spline interpolation (``Spline'').}
\label{fig:exp3}
\vspace{-6pt}
\end{figure}

Figure~\ref{fig:exp3} shows the results.  For the Alibaba dataset, no-filling
has a lower TPR than both forward filling and cubic spline interpolation,
implying that data filling is necessary for high prediction accuracy.  Also,
cubic spline interpolation achieves the highest TPR (92.8\% for A1 and 68.4\%
for B1).  A surprising observation is that linear interpolation has the lowest
TPR (even lower than no-filling).  One possible reason is that most
failure-related attributes series are of non-linear patterns (e.g., SM5,
SM187, and SM197).  For non-linear patterns, both forward filling and linear
interpolation fill in less accurate values compared to cubic spline
interpolation; even worse, linear interpolation may introduce more noise
samples than forward filling, as the latter only fills any missing gap with
duplicate but non-noise samples.

For the Backblaze dataset, cubic spline interpolation still achieves the
highest TPR (72.3\% for H1 and 82.4\% for S1).  Note that the rankings of
prediction accuracy across different filling methods are different in the
Alibaba and Backblaze datasets.  One possible reason is that the two datasets
have different data missing rates and patterns (Table~\ref{tab:missing} vs.
Table~\ref{tab:missing_bb}).

\subsection{Analysis of Whole \sysname}
\label{subsec:results2}

We evaluate \sysname as a whole with all data preprocessing techniques
enabled (i.e., failure-type filtering (for the Alibaba dataset only),
spline-based data filling, and automated pre-failure backtracking with the
observation window).  We compare \sysname with the {\em baseline}, in which we
only perform data filling with cubic spline interpolation on the raw dataset
but do not apply other data preprocessing techniques. 

\paragraph{Exp\#4 (Comparisons with state-of-the-arts).} We compare the
accuracy of \sysname with two state-of-the-art ML-based disk failure
prediction approaches \cite{botezatu2016,mahdisoltani2017}.  The ML model in
\cite{botezatu2016} is Regularized Greedy Forests (RGF) \cite{rie2014}, while
that in \cite{mahdisoltani2017} is Random Forests (RF) \cite{breiman2001}.
Our goal here is not to advocate any particular ML model; instead, we show
that \sysname improves the respective accuracy of a given ML model as a
general framework compared to the baseline. 


One challenge is that both approaches in \cite{botezatu2016,mahdisoltani2017}
build on completely different datasets, so it is infeasible to completely
reproduce their methodologies.  Thus, we make the ``best efforts'' to
implement their approaches based on their description and conduct our
comparison study under fair conditions.  Specifically, we train and test all
approaches using our datasets.  Since the feature construction process largely
depends on the data preprocessing and statistical analysis of the specific
datasets, for fair comparisons, we choose our constructed features for all
approaches.  

To this end, we compare the baseline and \sysname on three ML models:
LightGBM, RGF, and RF (i.e., six variants in total).  We implement the
baseline RGF and RF to resemble the approaches in \cite{botezatu2016} and
\cite{mahdisoltani2017}, respectively.  To alleviate data
imbalance, we perform the standard under-sampling technique on negative
samples as described in \cite{botezatu2016,mahdisoltani2017}, such that the
ratio of positive to negative samples is 1:10 (the prediction accuracy is
maintained for various ratios from 1:1 to 1:10 \cite{mahdisoltani2017}). In
contrast, for the Alibaba (Backblaze) dataset, \sysname and its baseline
implementation use the positive samples in the whole 10-month (3-month)
training phase and the negative samples on the last day observed
(\S\ref{subsec:evalmeth}).  Furthermore, we
conduct grid-search and configure the following model settings that achieve high
prediction accuracy: for LightGBM, we set 2,000 trees; for RGF, we set the
maximum number of trees as 2,000 and use L2-norm regularization for RGF; for RF,
we set 2,000 trees.  



\begin{table}[t]
\centering
\setlength\tabcolsep{3pt}
\begin{tabular}{|c|c|c|c|c|c|}
\hline
\textbf{ML models} &  & \textbf{A1} & \textbf{B1} &
\textbf{H1} & \textbf{S1} \\
\hline
\hline
\multirow{4}{*}{LightGBM} & 
\multirow{2}{*}{Baseline} &
\multirow{2}{*}{69.7\%} & \multirow{2}{*}{57.1\%} & 64.6\% & 76.2\% \\
 & & & & ($\pm$21.3\%) & ($\pm$4.4\%) \\
\cline{2-6}
& \multirow{2}{*}{\sysname}
& \multirow{2}{*}{\textbf{92.8\%}} &
\multirow{2}{*}{\textbf{68.4\%}} & \textbf{72.3\%} & \textbf{82.4\%} \\
 & & & & \textbf{($\pm$15.3\%)} & \textbf{($\pm$3.6\%)} \\
\hline
\multirow{4}{*}{RGF} & 
\multirow{2}{*}{Baseline} &
\multirow{2}{*}{67.2\%} & \multirow{2}{*}{53.4\%} & 56.4\% & 65.0\% \\
& & & & ($\pm$20.6\%) & ($\pm$8.1\%) \\
\cline{2-6}
& \multirow{2}{*}{\sysname}
& \multirow{2}{*}{\textbf{88.6\%}} &
\multirow{2}{*}{\textbf{67.2\%}} & \textbf{62.6\%} & \textbf{75.4\%} \\
& & & & \textbf{($\pm$17.3\%)} & \textbf{($\pm$5.0\%)} \\
\hline
\multirow{4}{*}{RF} & 
\multirow{2}{*}{Baseline} &
\multirow{2}{*}{62.6\%} & \multirow{2}{*}{48.3\%} & 50.7\% & 73.5\% \\
& & & & ($\pm$19.7\%) & ($\pm$3.4\%)\\
\cline{2-6}
& \multirow{2}{*}{\sysname}
& \multirow{2}{*}{\textbf{86.4\%}} &
\multirow{2}{*}{\textbf{64.5\%}} & \textbf{60.7\%} & \textbf{80.9\%} \\
& & & & \textbf{($\pm$17.6\%)} & \textbf{($\pm$3.5\%)} \\
\hline
\end{tabular}
\vspace{-3pt}
\caption{Exp\#4 (Comparison with state-of-the-arts). For H1 and S1, we also
report 95\% confidence intervals (in brackets) of all runs.}
\label{table:algorithms_comparison}
\vspace{-6pt}
\end{table}

Table~\ref{table:algorithms_comparison} shows the results.  \sysname increases
the TPR of all ML models over the baseline by 21.4-23.7\%, 11.3-15.8\%,
6.2-10.0\%, and 6.2-10.4\% for A1, B1, H1, and S1, respectively.  This shows
that \sysname improves the prediction accuracy for different ML models and
different datasets. 

\begin{figure}[!t]
\centering
\begin{tabular}{@{\ }c@{\ }c}
\includegraphics[width=1.65in]{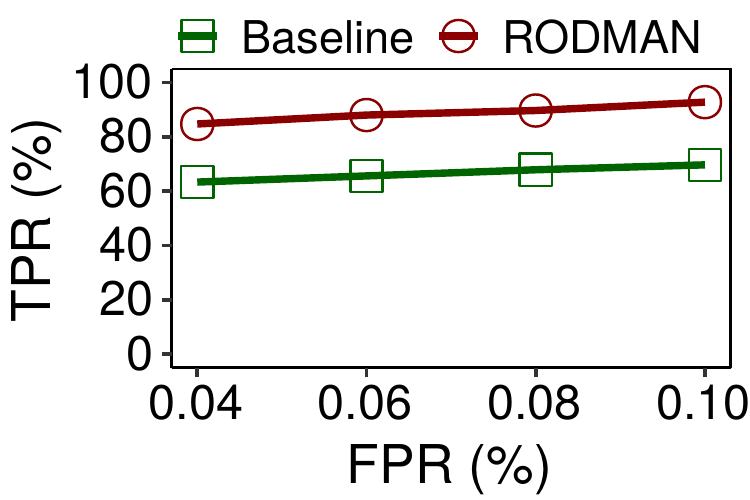} & 
\includegraphics[width=1.65in]{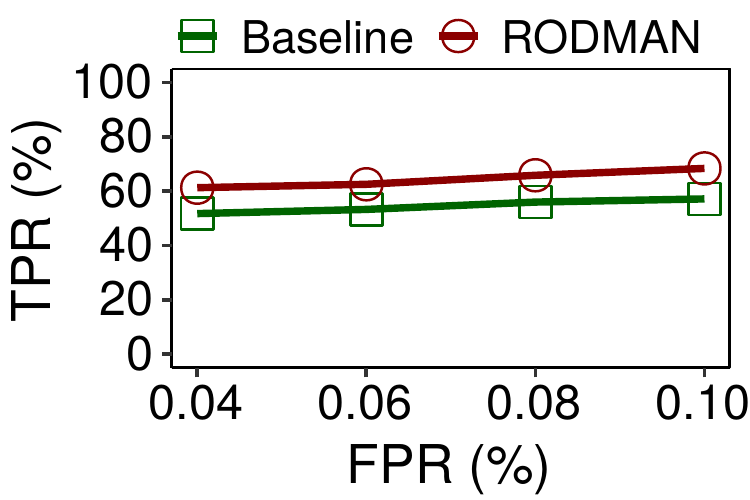} 
\vspace{-3pt}\\
\mbox{\small (a) A1} &
\mbox{\small (b) B1} 
\vspace{3pt}\\
\includegraphics[width=1.65in]{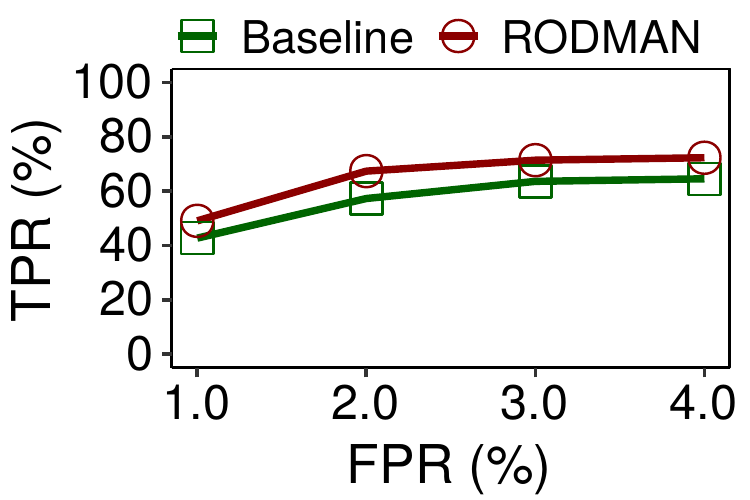} &
\includegraphics[width=1.65in]{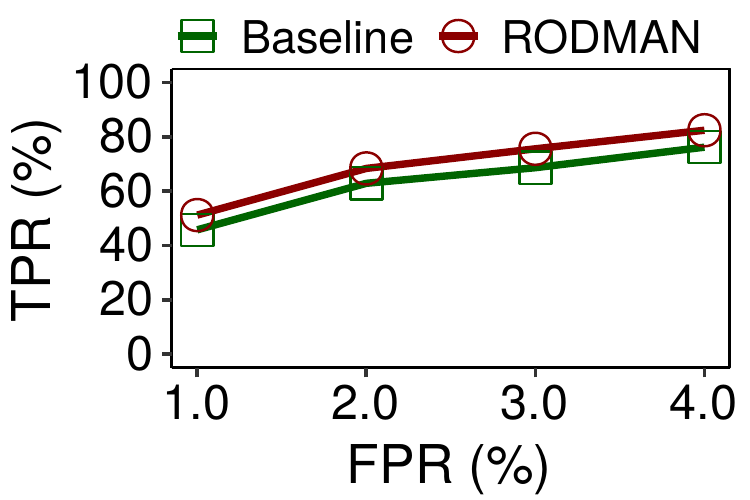} 
\vspace{-3pt}\\
\mbox{\small (c) H1} &
\mbox{\small (d) S1} 
\end{tabular}
\caption{Exp\#5 (Impact of FPR thresholds).}
\label{fig:exp4}
\end{figure}

\paragraph{Exp\#5 (Impact of FPR thresholds).}  In general ML-based
prediction, we can increase the TPR by increasing the FPR threshold (or vice
versa).  Here, we vary different FPR thresholds and measure the corresponding
TPR results to verify whether the prediction accuracy gain of \sysname over
the baseline (without \sysname) is still preserved in both Alibaba and
Backblaze datasets. 
Figures~\ref{fig:exp4}(a) and \ref{fig:exp4}(b) show the TPR results of A1
and B1, respectively, when we vary the FPR threshold from 0.04\% to 0.1\%,
while Figures~\ref{fig:exp4}(c) and \ref{fig:exp4}(d) show the TPR results of
H1 and S1, respectively, when we vary the FPR threshold from 1.0\% to 4.0\%.
The TPR gain of \sysname over the baseline is fairly stable for different FPR
thresholds, with around 20\%, 10\%, 8\%, and 6\% gains for A1, B1, H1, and S1
respectively. 

\begin{figure}[!t]
\centering
\begin{tabular}{@{\ }c@{\ }c}
\includegraphics[width=1.65in]{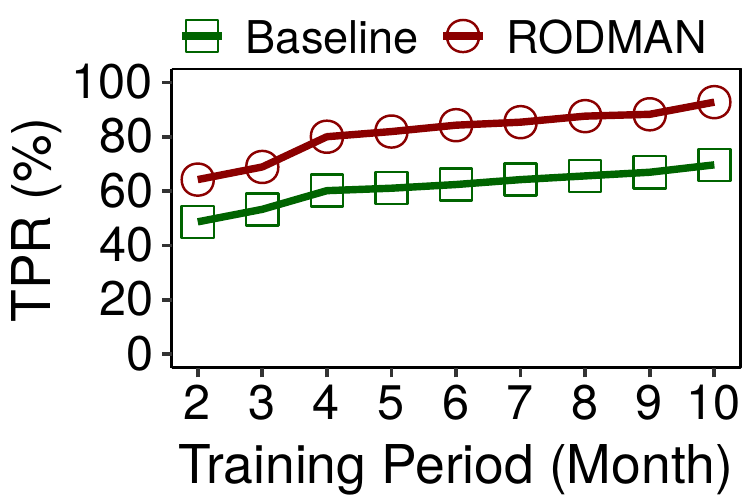} &
\includegraphics[width=1.65in]{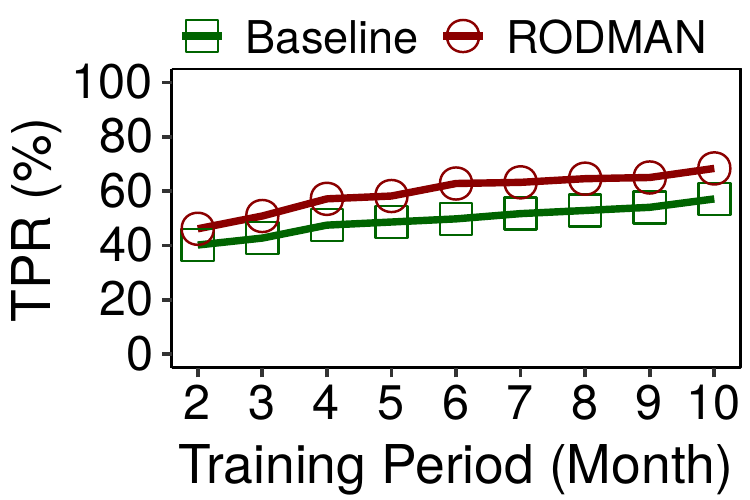} 
\vspace{-3pt}\\
\mbox{\small (a) A1} & 
\mbox{\small (a) B1}
\end{tabular}
\caption{Exp\#6 (Impact of training periods).}
\label{fig:training_period}
\end{figure}

\paragraph{Exp\#6 (Impact of training periods).}  Recall that for the Alibaba
dataset, we use a 10-month training period by default.  We now reduce the
length of the training period for positive samples.  Clearly, reducing the
length of the training period reduces the TPR, yet our goal here is to
evaluate the robustness of \sysname to the prediction accuracy gain over the
baseline (recall that we still choose one day of negative samples
(\S\ref{subsec:evalmeth})).  We vary the training period from 2 to 10 months,
counted backwards from April 2018, while we still set the test phase in May
2018.  Figure~\ref{fig:training_period} shows the impact of different training
periods.  If we reduce the training period for positive samples from the
default 10 months to 2 months, the TPR of \sysname drops from 92.8\% to 64.3\%
for A1 and from 68.4\% to 46.2\% for B1, since we now have fewer positive
samples for training.  Similar to Exp\#5, \sysname consistently increases the
TPR over the baseline by around 20\% and 10\% for A1 and B1, respectively, for
different training periods. 

\paragraph{Remarks.}
We show that \sysname improves the prediction accuracy via its data
preprocessing techniques, and its accuracy gain is shown in both Alibaba and
Backblaze datasets, different ML models, as well as different parameter
choices.
For the Alibaba dataset, we also validate the effectiveness of \sysname for
other disk models.  Here, we choose the disk models A2 and B2, since both of
them also have large numbers of disks, as well as large numbers of positive
samples for training.  The TPR of \sysname achieves 96.4\% and 63.2\% for A2
and B2, respectively, while that of the baseline only achieves 70.6\% and
43.9\%, respectively (we do not illustrate the results in plots here). 
We do not validate \sysname for other disk models (e.g., for vendors `C',
`D', and `E'), as we do not have sufficient positive samples for training and
our prediction analysis becomes less representative.  We pose the analysis
for such disk models in future work. 


\section{Related Work}
\label{sec:related}

\noindent
{\bf Field measurements.}  Extensive field measurement studies show the
prevalence of disk failures in production environments.  Earlier studies
\cite{schroeder2007,pinheiro2007} show that the actual disk failure rates are
much higher than the datasheet values.  In particular, Google's study
\cite{pinheiro2007} shows that over half of failed disks do not record any
SMART failure signal.  In addition to disk failures, faulty interconnects and
protocol stack errors are also dominant \cite{jiang2008}.  Furthermore, field
studies on NetApp clusters reveal patterns of latent sector errors
\cite{bairavasundaram2007,schroeder2010} and silent data corruptions
\cite{bairavasundaram2008}.  A more recent study from EMC \cite{ma2015}
analyzes the failures of about one million of disks over a five-year span, and
shows that reallocation sector counts are good indicators of predicting disk
failures.  Instead of emphasizing the prevalence of disk failures, our field
study focuses on motivating the need of data preprocessing. 
To address the limitations of SMART logs alone \cite{pinheiro2007}, we
leverage syslogs and trouble tickets for more accurate characterization
of disk failures. 



\paragraph{Disk failure prediction.}  A spate of studies show that highly
accurate disk failure prediction is achievable using classical statistical
techniques and machine learning models.  Examples include Bayesian classifiers
\cite{hamerly2001}, rank-sum tests \cite{hughes2002}, Markov models
\cite{eckart2009,zhao2010}, rule-based learning \cite{agarwal2009},
back-propagation neural networks \cite{zhu2013} and recurrent neural networks
\cite{li2014}, regularized greedy forests \cite{botezatu2016}, random forests
\cite{mahdisoltani2017}, and online random forests \cite{xiao2018}.  Based on
disk failure prediction, some studies further propose proactive fault
tolerance approaches to protect against soon-to-fail disks, such as proactive
replication \cite{li2016procode} and adaptive scrubbing
\cite{mahdisoltani2017}.  However, existing studies mainly validate their
proposals using the datasets from a small-scale disk population (e.g., 
\cite{hamerly2001,hughes2002,eckart2009,zhao2010,agarwal2009,zhu2013,li2014,li2016being,li2016procode})
or the public Backblaze SMART logs \cite{backblaze} (e.g.,
\cite{botezatu2016,mahdisoltani2017,xiao2018}).  We complement these studies
through our proposed data preprocessing techniques and the correlation of
multiple data sources (i.e., SMART logs, syslogs, and trouble tickets). 


\section{Conclusion}

We have witnessed extensive ML-based disk failure prediction studies in the
literature, but limited emphasis is put on proper preprocessing on training
datasets.  This paper fills the void and posits the need for improving the
quality of the training datasets used for ML model training instead of
designing new ML models.  We present \sysname, a robust disk failure
prediction management pipeline that is now deployed in a large-scale
cloud service provider. It builds on three data preprocessing techniques
(i.e., failure-type filtering, spline-based data filling, and automated
 pre-failure backtracking) to carefully refine and correlate SMART logs,
syslogs, and trouble tickets before feeding them into any ML model.  We show
that \sysname improves the disk failure prediction accuracy for general ML
models on different datasets.

\bibliographystyle{abbrv}
\bibliography{bibliography}

\end{document}